\documentclass[10pt]{article}

\usepackage[margin=0.95in]{geometry}
\usepackage{graphicx}
\usepackage{amsmath}
\usepackage{amssymb}
\usepackage{booktabs}
\usepackage{natbib}
\usepackage{hyperref}
\usepackage{xcolor}
\usepackage[compact]{titlesec}

\titlespacing*{\section}{0pt}{1.4ex plus 0.5ex minus 0.2ex}{0.8ex plus 0.2ex}
\titlespacing*{\subsection}{0pt}{1.1ex plus 0.4ex minus 0.2ex}{0.6ex plus 0.2ex}
\titlespacing*{\paragraph}{0pt}{1.0ex plus 0.3ex minus 0.2ex}{0.6em}

\hypersetup{
  colorlinks=true,
  linkcolor=blue,
  citecolor=blue,
  urlcolor=blue
}

\newcommand{\reasoncode}{\texttt{C0\_\allowbreak INSTRUMENT\_\allowbreak INVALID}}

\title{An Exploratory Replica-Overlap Probe of the Grokking Transition}

\author{%
  A. C. Opus, J. Q. Lu\thanks{Correspondence: \texttt{junqiang.lu@upr.edu}.}\\
  \small Department of Physics, University of Puerto Rico, Mayag\"uez, PR 00680, USA
}

\date{2026-09-21}

\begin{document}
\maketitle

\begin{abstract}
We trained 64 independently seeded networks in four configurations, continuing each to
sustained convergence or a 40{,}000-epoch ceiling. We then asked whether an RSB-inspired
distribution of pairwise weight overlaps changes across the grokking transition.
\textbf{It is the alignment step, not the overlap statistic, that determines what this
registered probe can report.} The registered implementation permutes hidden units
without the corresponding bias and head-internal permutations and therefore does not
preserve the network function. Every $q_{wt}$ value computed through this alignment
inherits the defect; $q_{fn}$ does not, because it is computed from predictions of the
unpermuted models. The numerical-precision requirement also failed, and an audit found
protocol deviations. Consequently, the pre-registered rule gives no verdict:
\textbf{registered outcome: \texttt{UNDETERMINED} (reason code: \reasoncode{})}. These
data provide neither a confirmatory null nor a validated reading of the Parisi order
parameter. Only \texttt{frac40} cleared the 12/16 checkpoint-completeness requirement.
For this configuration, a post-hoc criterion applied to the same data gave a Hartigan-dip
interval containing zero (95\% CI for $\Delta\text{dip} = [-0.017, 0.034]$), whereas the
overlap standard deviation increased by a factor of about $5.6$. A post-hoc calibration
assigns the dip test zero power at the simulated separations; the interval is therefore
uninformative, not evidence of no change. The standard-deviation ratio is the only
statistic here with power at the observed effect. Ensemble loss was near-flat only under
the pre-specified $1\%$ threshold. Finally, grokking rates of $0/16$, $11/16$ and $16/16$
remain descriptive because train fraction is confounded with split identity.
\end{abstract}

\section{Introduction}

Grokking occurs when a network's held-out accuracy jumps from chance to near-perfect
performance long after its training loss has saturated \citep{power2022grokking}. It is
widely described as a phase transition, and recent work treats it explicitly as one.
\citet{nanda2023progress} identify Fourier-basis ``progress measures'' that anticipate
the transition mechanistically. Two contemporaneous 2026 papers instead use
statistical-mechanical descriptions: \citet{cullen2026basin} employ the Local Learning
Coefficient from Singular Learning Theory, whereas \citet{xu2026earlywarning} employ a
gradient-``commutator defect'' as an early-warning signal. Direct inspection shows that
both are \emph{single-network, single-trajectory} geometric quantities. Neither measures
agreement or disagreement among several solutions trained independently on the same
task.

Statistical physics provides a tool made for this multi-solution question: the replica
method, and in particular Parisi's replica-symmetry-breaking (RSB) solution of the
Sherrington-Kirkpatrick spin-glass model \citep{sherrington1975solvable,
parisi1980order}. For $n$ independent replicas of a disordered system sharing the same
quenched disorder, RSB theory predicts the form of the pairwise \emph{overlap}
distribution $P(q)$, $q^{ab} = \tfrac{1}{N}\sum_i s_i^a s_i^b$. The replica-symmetric
(RS) phase gives a single delta function at one value; when RSB sets in, the distribution
becomes nontrivial and may be multi-modal \citep{mezard1987spin}. RSB methods have
already been applied to neural networks: to perceptron storage capacity
\citep{gardner1988optimal}, dense associative memories at scale
\citep{albanese2021replica}, Gibbs samples from a Restricted Boltzmann Machine
\citep{hartnett2018replica}, trained restricted Boltzmann machines directly
\citep{fachechi2024replica}, and feedforward networks mapped onto spin models and
followed through training \citep{barney2024spin}. However, to our knowledge, and as
confirmed by the live literature search in Section~\ref{sec:related}, these methods have
not been used to ask whether the shape of $P(q)$ changes \emph{across the grokking
transition itself} for genuinely independent replicas rather than for the local geometry
of one trajectory.

In this paper, we report a pre-registered attempt to answer that question. The
registered statistical test never clears its instrument-validity gate
(Section~\ref{sec:o1}), so the study does not yield a confirmatory null. On the only
configuration eligible for an overlap comparison, the exploratory analysis finds no
evidence that the shape of the distribution changes in an RSB-consistent way between the
memorization plateau and the post-grokking state. We report that result together with a
second observation, not pre-registered, allowed by the same 64-replica sweep: grokking
frequency is descriptively associated with train fraction across three
fraction$\times$split combinations in which fraction and split identity are confounded
(Section~\ref{sec:reliability}).

\section{Related Work}
\label{sec:related}

\paragraph{Spin-glass constructions applied to neural loss landscapes.}
\citet{bae2026dataset} adapt the Franz--Parisi construction to finite networks. Using
adaptive sequential Monte Carlo, they measure local entropy---the effective volume of
low-loss configurations at each distance from a reference solution---and relate it to
dataset complexity. This is the published method closest to the order parameter used
here, but the objects remain distinct. Their object is an overlap-\emph{constrained}
volume around a single reference, evaluated on trained solutions; ours is a distribution
of pairwise overlaps between independently trained replicas, evaluated across a
transition. Thus, neither study answers the other's question: they do not study
grokking, and we do not measure local entropy.

\paragraph{Perturbation probes of the grokking plateau.}
\citet{lin2026canalization} apply short weight-decay pulses to the pre-generalization
plateau. They report a stable dose ordering in the resulting shift of generalization
time, together with test-loss barriers between perturbed and baseline checkpoints that
\emph{collapse toward zero}. The barrier result bears on the alignment premise used here
(Section~\ref{sec:method}), but it does not settle that premise. Their barriers join two
checkpoints from one training trajectory, one perturbed and one not. Our premise concerns
\emph{independently initialised} replicas compared after permutation alignment. The two
quantities differ. Moreover, this study cannot make the comparison directly because its
registered alignment cross-check was among the controls that did not run
(Section~\ref{sec:deviations}, D3). Running that check against their setup is the clear
next measurement; we flag it without treating their result as support.

\paragraph{Grokking as a phase transition.} \citet{power2022grokking} define the
phenomenon and the task family inherited here: small algorithmic datasets and modular
arithmetic. \citet{nanda2023progress} give the first mechanistic account for modular
addition. Two lines of work map grokking directly onto phase-transition theory.
\citet{zunkovic2022grokking} derive exact critical exponents and grokking-time
distributions for solvable rule-learning models through a tensor-network map to
perceptron statistical learning theory. By contrast, \citet{rubin2023grokking} use an
adaptive-kernel feature-learning theory to identify the post-grokking state as analogous
to the mixed phase following a first-order transition. \citet{cullen2026basin} and
\citet{xu2026earlywarning} instead describe grokking as a basin-selection or geometric
phase transition. Direct content checks confirm that their measures concern one network
and one trajectory; neither paper contains a cross-replica overlap or RSB order parameter.
Nor do the two phase-transition-theory studies use a replica overlap distribution: their
stated methods are based on kernels and tensor networks, not replicas.

Two 2026 additions sharpen the same boundary from outside the replica picture.
\citet{kataria2026quantifying} map the memorization-to-generalization boundary across
384 configurations of two-hidden-layer MLPs on modular arithmetic. They fit the onset-time
power law
($T_{\mathrm{grok}} \propto H^{-0.27} D^{-2.04} \eta^{-0.50} \lambda^{-0.64}$,
$R^2 = 0.732$) and locate a sharp weight-decay boundary at $\lambda \gtrsim 1.0$
separating grokking from non-grokking configurations. \citet{wang2026generalization}
derive a sufficient condition for delayed generalization by bounding a
prediction-variation term along the training trajectory. Both are single-network
accounts: the first tracks weight \emph{norm}, whereas the second tracks a function-space
oscillation bound. Neither forms an overlap between replicas. The weight-decay boundary
is nevertheless the hyperparameter-space counterpart of the non-grokking control arm
used here; this study does not perform an external check on it.

\paragraph{Multi-replica structure without RSB language.} Other work compares
\emph{multiple} independently trained networks without invoking replica theory.
Mode-connectivity studies \citep{garipov2018loss,draxler2018essentially} find that simple
low-loss paths connect independently trained optima. Permutation-alignment work
\citep{ainsworth2022git,entezari2021role} argues that this connectivity appears after
permutation symmetry has been quotiented out. Accordingly, this study adopts the Git
Re-Basin algorithm of \citet{ainsworth2022git} for alignment
(Section~\ref{sec:method}). \citet{liao2024exploring} apply a spin-glass view to DNN loss
landscapes and report a hierarchy among trained solutions reminiscent of RSB, although
they obtain it by hierarchical clustering rather than from an overlap distribution.
\citet{sharma2024simultaneous} complicate the single-basin picture: pairwise (``weak'')
alignment does not imply one permutation that aligns all networks simultaneously
(``strong'' alignment). This suggests structure beyond a trivial RS picture. The
lottery-ticket literature \citep{frankle2019lottery,frankle2020linear} independently
motivates the ``ensemble size'' half of the present design: a large, redundant family of
near-equivalent solutions within one trained network's solution space.

\paragraph{Explicitly RSB-adjacent 2025--2026 work.} Three recent studies approach this
question, but construct their replicas differently. \citet{zhang2025grokking} (NeurIPS
2025) cast grokking as glass relaxation through Wang-Landau sampling of a
Boltzmann-entropy landscape. Direct inspection confirms that the analysis remains
single-trajectory, with no cross-replica overlap or RSB/Parisi order parameter.
\citet{li2025spinglass} is the closest paper found in this survey: it uses
RSB/Parisi-adjacent vocabulary and refers directly to grokking (``$Q^{ab}$ curves reveal
structural changes\dots{} after loss saturation''). However, its meaning of ``replica''
differs from the one required here. Direct inspection of the method shows that its
replicas are Gibbs/thermal samples of a Hopfield model derived from \emph{one}
already-trained network's weights (disorder $=$ the frozen trained weights; replica $=$ a
thermal sample of that single derived system). They are not network instances trained
independently under shared quenched disorder, the meaning used in the present
RSB-inspired design. \citet{chan2026tunneling} introduce a metric called ``replica
correlation,'' but branch their replicas from a shared checkpoint rather than training
them independently from scratch. Again, this is different from independent draws under
the same quenched disorder. Clearly, none of the three studies combines the elements of
this design: independently seeded replicas, a full distribution of pairwise overlaps,
and a registered decision rule tied to the grokking transition itself.

\section{Method}
\label{sec:method}

\paragraph{Task and architecture.} We use the modular-arithmetic addition task of
\citet{power2022grokking} with modulus $p = 113$ and the architecture pinned in the
pre-registration for this study (\S B2). We ran four configurations: \texttt{frac25},
\texttt{frac30}, and \texttt{frac40}, with train\_frac $\in
\{0.25, 0.30, 0.40\}$, and the manipulation control \texttt{nogrok}. The control uses
i.i.d.\ random labels at train\_frac $= 0.30$ and the same split seed as
\texttt{frac30}; it is therefore a matched-split control and is expected never to
generalize. Each configuration contains 16 independently seeded replicas. Training
stopped when test accuracy $\geq 99\%$ had remained there for 1{,}000 consecutive epochs,
or otherwise at a 40{,}000-epoch ceiling.

The stopping condition differs across configurations and determines how their
checkpoints must be read. All 16 \texttt{frac40} replicas stopped early, at final epochs
$4{,}372$--$26{,}838$. The \texttt{frac30} outcome is mixed: $11/16$ stopped early,
whereas the $5$ replicas that never grokked exhausted the budget at epoch $39{,}999$, the
last index of the zero-based $40{,}000$-epoch budget. All 16 replicas in both
\texttt{frac25} and \texttt{nogrok} likewise exhausted the ceiling without meeting the
criterion. Thus, $37$ of the $64$ runs reached the ceiling, but none belongs to
\texttt{frac40}, the only configuration analysed for overlap. Its checkpoints are
\emph{event-aligned}: they were taken a fixed number of epochs after the convergence
criterion fired, not at one common absolute epoch. No run was trained deep into a
post-transition regime. All 64 runs completed without a crash ($\approx$3.4h wall clock,
4-way parallel, local M3 Ultra, \$0 cloud spend).

\paragraph{Pre-registered gate (C1).}\begin{sloppypar} Before any overlap analysis, a
configuration must clear a \emph{checkpoint-completeness} gate: at least 12 of its 16
replicas must reach both a memorization-plateau checkpoint and a post-grokking checkpoint
within the epoch budget. This is an eligibility criterion. It asks whether enough saved
checkpoints exist for a comparison; it is \emph{not} a power calculation, and no such
calculation was run (Section~\ref{sec:deviations}, D4). A configuration that misses the
gate is reported as \texttt{RAISE}, meaning ineligible for overlap analysis, rather than
being folded into a null verdict on RSB.
\end{sloppypar}

\paragraph{Alignment, and a defect in it.} Pairwise overlap is meaningful only after
arbitrary permutation symmetry has been quotiented out. Within each configuration, we
align every pair by Git Re-Basin-style weight matching \citep{ainsworth2022git} over two
permutation groups: the 128-dimensional residual stream and the 512-dimensional MLP
hidden layer.

\textbf{The implementation does not realise a function-preserving permutation, and we
report the analysis knowing this.} A post-hoc audit of \texttt{pilot/align.py} found that
\texttt{apply\_permutation()} permutes only nine \texttt{*.weight} tensors. Yet every
linear layer has a bias, and those biases are nonzero in every checkpoint analysed.
Across all sixteen \texttt{frac40} post-grokking replicas, every component of
\texttt{mlp\_in.bias}, \texttt{W\_O.bias} and \texttt{mlp\_out.bias} is nonzero; the
median $\max_i|b_i|$ values are $0.382$, $0.084$ and $0.020$, respectively. Thus, the
permutation changes the function instead of preserving it. Specifically,
\texttt{mlp\_in.bias} is not permuted with $\pi_{\mathrm{mlp}}$, while
\texttt{W\_O.bias} and \texttt{mlp\_out.bias} are not permuted with
$\pi_{\mathrm{resid}}$; attention head-internal symmetries are not handled at all. The
permuted network is consequently not the same function as the original. The required
symmetry has not, in fact, been quotiented out.

Why did the original positive control miss the defect? The synthetic twin was scrambled
and restored with the \emph{same} incomplete \texttt{apply\_permutation()}. Recovering
overlap $1.000000$ (vs.\ $-0.010$ unaligned) therefore shows only that the array operation
is invertible; it never tests whether the twin and the original produce identical logits.
The discriminating result for two genuinely different replicas ($0.182$ aligned vs.\
$-0.001$ unaligned) does not resolve the problem either. Likewise, the fp32/fp64 agreement
check reported below establishes numerical repeatability of the permutation search, not
its correctness. Alignment quality may itself vary between checkpoints, so this defect
can create or erase structure in $P(q)$. It is the principal reason we report the
instrument as invalid for the registered question.

\paragraph{Overlap statistic and decision rule.} The primary statistic is $q_{wt}$, the
pairwise weight-space overlap after alignment. During pre-registration review, this
measure replaced the originally drafted function-space statistic after a union-bound
argument in the pre-registration showed that statistic to be $\geq 0.98$ for every
post-grokking pair as a mathematical consequence of the $99\%$ accuracy threshold alone;
no pilot data existed at that point. \textbf{This replacement is not a protocol
deviation.} The pre-registration committed on 2026-08-22 at 18:15 names $q_{wt}$ as the
primary statistic in \S7\footnote{Section references of the form \S7, \S8, \S13 and \S B2 denote numbered sections of \emph{the pre-registration}, not of this paper. The pre-registration is \path{pre_registered_hypotheses/20260822_P006_rsb_grokking.md}; its body was committed as \texttt{bd69ea4} on 2026-08-22 at 18:15, before any replica finished training, and two later commits (\texttt{bd0d32e} at 06:54 and \texttt{ce5a859} at 10:52 on 2026-08-23) amend it after the data existed.}. The replicas did not finish training until 2026-08-23 at 00:10,
and overlap results were not produced until 02:59. The replacement therefore predates
the data to which it was applied and is registered, unlike the \S13 criterion discussed
below. The pre-registration left \emph{both} directions open: $P(q_{wt})$ could acquire
structure across the transition, becoming wider or more multi-modal after grokking, or
its structure could collapse. We registered no one-sided prediction and report none.

We quantify shape change with Hartigan's dip statistic, implemented in the
\texttt{diptest} package and validated on synthetic unimodal/bimodal data. A percentile
bootstrap then tests whether $\Delta\text{dip} = \text{dip}_{\text{post}}
- \text{dip}_{\text{plateau}}$ excludes zero. This procedure was chosen over a $z$-score
gate because the bootstrap distribution of the dip statistic is non-normal and
boundary-truncated. As a precondition for trusting $\Delta$dip, a registered
numerical-precision rail requires fp32/fp64 agreement within $10^{-6}$ for each pair. If
the rail fails, the result is \textbf{registered outcome: \texttt{UNDETERMINED} (reason
code: \reasoncode{})}; the $\Delta$dip value is not treated as meaningful.

\paragraph{The function-space measure $q_{fn}$, and why it was replaced.} For a pair of
replicas, $q_{fn}$ is the fraction of the shared held-out test set on which their
\emph{top-1 predictions} agree. This is exact $\arg\max$ label agreement over the
$7{,}661$ held-out $(a,b)$ pairs at train fraction $0.40$, not a probability or logit
distance and not thresholded confidence. Every \texttt{frac40} replica reaches
$\texttt{test\_acc} = 1.0000$, so all replicas agree with the ground truth, and thus with
one another, on every held-out item. This gives $q_{fn} = 1.0000$ (sd $0.0000$)
post-grokking, against $0.023$ at the plateau. It is this saturation that motivated the
change to $q_{wt}$. The saturation follows from the accuracy ceiling and this definition
of agreement; it is not a finding about overlap structure.

\paragraph{Ensemble-size measure (O2).} As a second, independent probe of the same
underlying question, we measure $k^*$, the smallest ensemble size---with probabilities
averaged over $k$ member replicas---whose loss is within 1\% of the full-ensemble loss.
The calculation exhaustively enumerates one pre-registered 8+8 replica split. The
pre-registration required exhaustive enumeration only through $k=4$ and a $100$-subset
bootstrap for $k>4$. With $n=8$, however, the largest exhaustive set is
$\binom{8}{4}=70$ subsets. We therefore enumerate every $k$, which is exact where the
registered bootstrap would have resampled, and record the substitution as deviation D7
rather than present it as the registered procedure. In an RSB-consistent picture,
diminishing returns from ensembling should track the modality of $P(q)$.

\section{Protocol deviations}
\label{sec:deviations}

This study departs from its pre-registration in several respects. We collect the
departures here because their combined effect, rather than any one of them, reduces the
study from a confirmatory test to an exploratory one.

\paragraph{D1. Checkpoints are not at the registered epochs.} The pre-registration
requires the \emph{first} epoch at which memorization onset and post-grokking convergence
are observed, together with an archive every 500 epochs. Instead, the training code
waits for the sustained criterion to finish and saves the weights current at that time.
Consequently, \texttt{memorization\_plateau.pt} is late by 499 epochs and
\texttt{post\_grokking.pt} by 999 epochs; the every-500-epoch archive was never
implemented. The reported comparison is not between the two registered states, and the
intermediate checkpoints needed to reconstruct the registered analysis do not exist.

\paragraph{D2. The analysed configuration stopped early, so ``deep in the grokked
phase'' does not apply to it.} A run stopped when its convergence criterion had held or,
if it never met that criterion, at the ceiling (Section~\ref{sec:method}). The sweep logs
give the following values for each configuration:

\begin{center}
\begin{tabular}{llll}
\hline
configuration & final epochs & mean & stopped early / hit ceiling \\
\hline
\texttt{frac40} & 4{,}372--26{,}838 & 11{,}525 & 16 / 0 \\
\texttt{frac30} & 31{,}063--39{,}999 & 37{,}001 & 11 / 5 \\
\texttt{frac25} & 39{,}999 (all) & 39{,}999 & 0 / 16 \\
\texttt{nogrok} & 39{,}999 (all) & 39{,}999 & 0 / 16 \\
\hline
\end{tabular}
\end{center}

Thus, $37$ of the $64$ runs reached the ceiling; the blanket statement in an earlier
draft that none did was wrong. The point relevant to the overlap analysis is narrower:
\emph{every} \texttt{frac40} replica---and this is the only configuration analysed---stopped
early, at a median well below one third of the budget. The data therefore do not support
the claim that these replicas remained deep in a post-transition phase for the rest of
training, and we have removed that claim.

\paragraph{D3. Registered controls that were not executed.} Several specified
procedures did not run: the 10\% Entezari alignment cross-check; the
alignment-convergence check; the saving and analysis of overlap drift for
\texttt{nogrok} and \texttt{frac30} at \emph{matched absolute epochs}; the O2 200-split
exploratory companion described in the pre-registration as ``retained, run, and
reported''; and \texttt{power\_calibration.py}, listed there as a mandatory pre-analysis
deliverable. No corresponding file or result exists in \texttt{pilot/}.

The consequence for \texttt{nogrok} is precise. This control was registered to exclude
the possibility that continued optimisation and weight decay move $P(q)$ by themselves.
Because its overlaps were never analysed at epochs matched to the grokking runs, it does
not perform that role here; it shows only that random labels do not generalise. Excluding
that alternative would require retraining, which this paper deliberately does not do.
We report the existing data and disclose the gap.

\paragraph{D4. The registered power calibration was not run before the analysis; a
post-hoc one was run afterwards, and it matters.} No pre-analysis power calibration was
performed. Accordingly, this paper makes no claim of statistical power by design. The
12/16 gate measures checkpoint completeness, not power. A post-hoc calibration was run
on 2026-09-02, after the results reported here. We disclose it because it changes how
the dip interval must be interpreted. Simulations of the registered dip statistic at
separations of $0.00$, $0.01$ and $0.02$ give power $0.00$ at every point. By contrast,
the overlap-standard-deviation ratio reaches power $0.758$ at a ratio of $1.5$ and
$1.000$ at the observed ratio of $5.56$, for a minimum detectable effect of about a
$1.5$-fold ratio. \textbf{The dip interval reported below is therefore uninformative,
not null-supporting: at the simulated separations, the test could not have detected a
change had one been present.} The calibration is itself provisional because it is scaled
from the route-A \texttt{frac40} values produced by the defective alignment. We report it
as a caveat on the dip result, not as a result. The information ceiling is 16 independent
replicas; their 120 pairs are strongly dependent. Moreover, no equivalence margin is
reported, so an interval containing zero cannot exclude an effect of practical size.

\paragraph{D5. The alignment implementation is defective.} As explained in
Section~\ref{sec:method}, permutations are applied to the weights but not to the
corresponding biases, and head-internal symmetries are left untreated. This departure
makes the instrument invalid for the registered question; it does not merely leave the
instrument under-powered.

\paragraph{D6. The \texttt{nogrok} labels are not the registered ones.} The
pre-registration specifies ``shuffled labels: a fixed random bijection-free permutation
of the $(a+b)\bmod p$ targets, drawn once per configuration''. The implementation instead
draws one i.i.d.\ uniform label for each $(a,b)$ pair, once, and shares it across all
sixteen replicas. This substitution is deliberate and documented in
\texttt{pilot/model.py}. A permutation $\sigma((a+b)\bmod p)$ remains a deterministic,
generalisable function of $(a+b)\bmod p$, learnable as the modular sum followed by a
$113$-entry lookup; it would therefore defeat the control's purpose of supplying no
generalisable structure. We judge the substitution to serve the registered intent better
than the registered wording, but it remains a deviation and is recorded as such.

\paragraph{D7. O2 uses exhaustive enumeration where a bootstrap was registered.} The
pre-registration calls for exhaustive enumeration of ensemble subsets when $k\leq4$ and
a $100$-subset bootstrap sample when $k>4$. Instead, the implementation enumerates every
$k$ exhaustively. With group size $n=8$, the largest subset count is
$\binom{8}{4}=70$, below the $100$ draws required by the registered bootstrap. The
substitution therefore makes the curve exact rather than resampled. The code includes an
assertion that fails if the count ever exceeds $100$, so a larger $n$ cannot silently
skip the registered branch. Nevertheless, this is a departure from the registered
procedure, and we record it as one.

\section{Results}

\subsection{Grokking reliability across train fraction}
\label{sec:reliability}

\begin{figure}[h]
\centering
\includegraphics[width=0.48\textwidth]{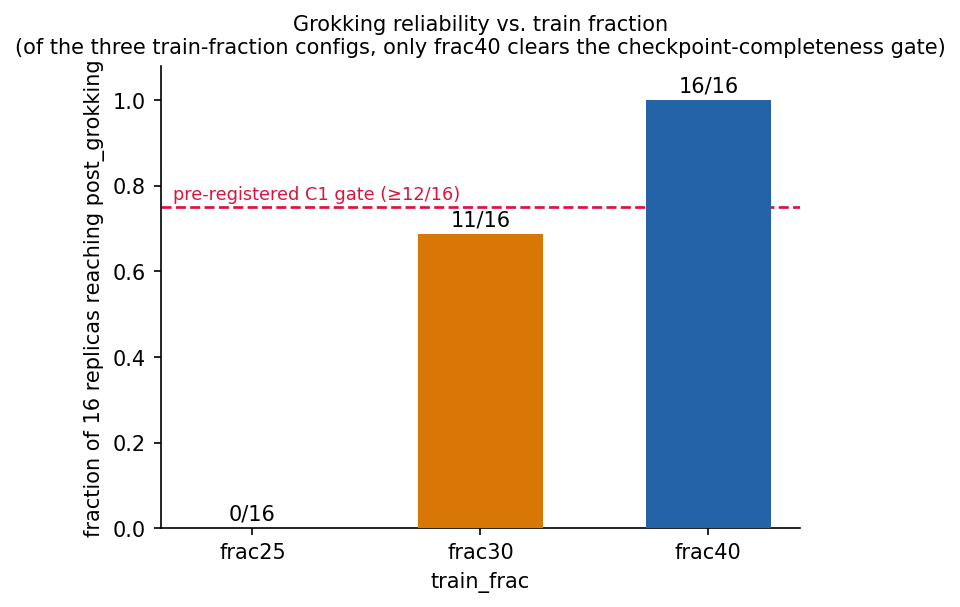}
\includegraphics[width=0.48\textwidth]{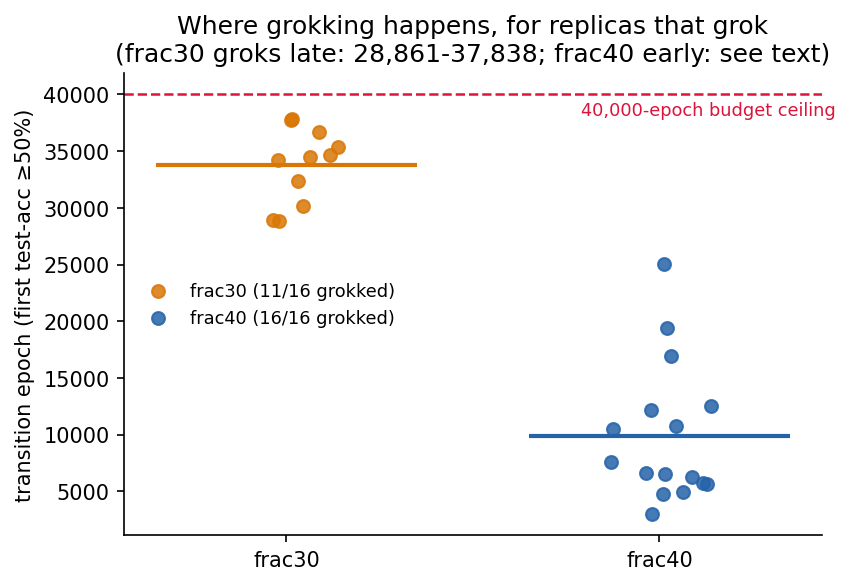}
\caption{Left: fraction of 16 replicas that reach post-grokking in each realised
\texttt{train\_frac}$\times$split configuration. \textbf{Train fraction is confounded
with split identity} (one split seed per fraction); the horizontal axis therefore
indexes three realised configurations rather than a fraction effect. Of the three
train-fraction configurations, only \texttt{frac40} clears the registered
checkpoint-completeness gate. Right: transition-epoch distributions for the two
configurations in which at least some replicas grok. The 11 grokked \texttt{frac30}
replicas transition at epochs $28{,}861$--$37{,}838$, a span of about $9{,}000$ epochs;
even the latest transition leaves $2{,}161$ epochs in the budget.}
\label{fig:reliability}
\end{figure}

Table~\ref{tab:c1} and Figure~\ref{fig:reliability} show the C1 gate outcome. This is the
first substantive result of the sweep and is independent of the RSB question.

\begin{table}[h]
\centering
\small
\small
\small
\begin{tabular}{lcccc}
\toprule
Config & Memoriz. & Post-grok. & Gate ($\geq$12/16) & Transition epoch \\
\midrule
\texttt{frac25} & 16/16 & 0/16 & RAISE & --- \\
\texttt{frac30} & 16/16 & 11/16 & RAISE (one short) & 28{,}861--37{,}838 (mean 33{,}765) \\
\texttt{frac40} & 16/16 & 16/16 & \textbf{PASS} & mean 9{,}899 \\
\texttt{nogrok} (control) & 16/16 & n/a by design & n/a & --- \\
\bottomrule
\end{tabular}
\caption{Grokking reliability for the three realised fraction-by-split configurations.
Train fraction is completely confounded with split identity because each fraction uses a
different split seed. The panel therefore compares the three realised configurations;
it does not estimate an effect of fraction. Among the three train-fraction
configurations, only \texttt{frac40} clears the registered checkpoint-completeness gate
for overlap analysis. The \texttt{frac25} and \texttt{frac30} configurations are
reported as \texttt{RAISE} (ineligible), not as null evidence about RSB. By design, the
\texttt{nogrok} control does not participate in this gate. No \texttt{frac25} replica
reaches a post-grokking checkpoint within budget, and \texttt{nogrok} is not expected to
do so; the control's final test accuracy averages $0.90\%$, compared with the
$1/113 \approx 0.885\%$ chance floor. For the replicas that grok, the transition column
reports the first epoch at which test accuracy is $\geq 50\%$.}
\label{tab:c1}
\end{table}

The \texttt{RAISE} branch behaved as registered: two configurations fell below 12/16 and
were withheld from overlap analysis rather than scored. We report that outcome instead
of working around it. Correct behavior of this branch should not be mistaken for overall
fidelity to the pre-registration, because several other registered components did not
run (Section~\ref{sec:deviations}). For \texttt{frac25}, 0/16 replicas grokked within the
budget. This is a statement about reachability, not evidence against RSB.
\texttt{frac30} misses the gate by one replica; its 11 grokked replicas transition at
epochs $28{,}861$--$37{,}838$, with a mean transition epoch of $33{,}765$ out of
$40{,}000$. We report this range instead of calling the distribution tightly clustered:
the span is about $9{,}000$ epochs, and the latest transition still leaves $2{,}161$
epochs in the budget. Thus, these data do not show the runs finishing at the edge. The
pattern nevertheless suggests that more of the five ungrokked replicas might have grokked under a larger
budget. Extending selected runs after the fact would violate the study's ordering of
registration before the final experiment, so we leave the limitation in place.

\begin{figure}[h]
\centering
\includegraphics[width=0.55\textwidth]{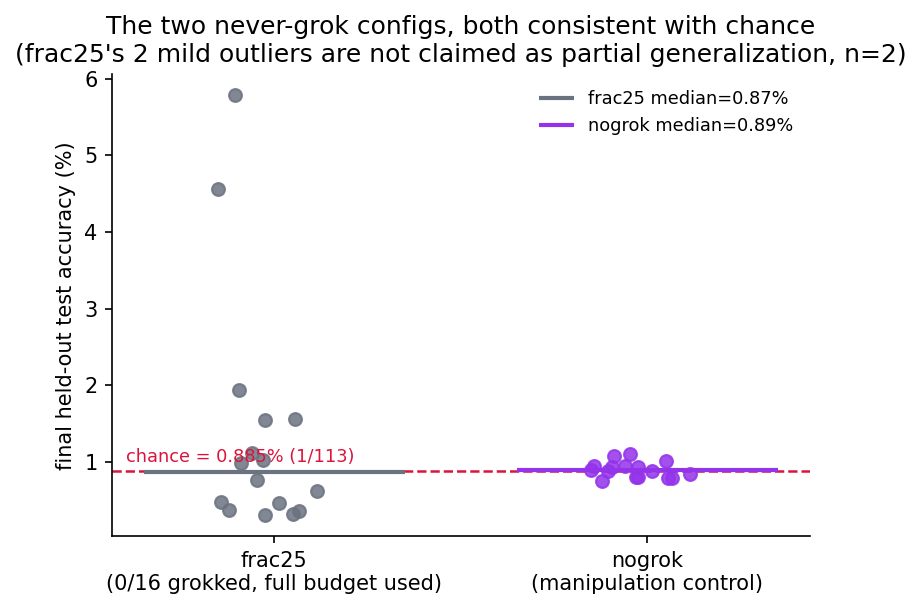}
\caption{Held-out test accuracy for the two configurations that never grok within the
budget. Both follow the $1/113$ chance floor. Two \texttt{frac25} outliers ($4.6\%$ and
$5.8\%$, compared with a \emph{median} of $0.87\%$ across the 16 replicas) are reported
only descriptively ($n=2$), not as evidence of partial generalization.}
\label{fig:reliability_test}
\end{figure}

Figure~\ref{fig:reliability_test} shows that the sample means and medians of held-out test
accuracy for \texttt{frac25} and \texttt{nogrok} both lie near the chance floor
($1/113 \approx 0.885\%$). This comparison is descriptive: we ran no test and specified
no equivalence margin. The pattern is the one expected if \texttt{frac25} fails to
generalise within budget rather than generalising imperfectly. Two replicas attain
$4.6\%$ and $5.8\%$, against a median of $0.87\%$; we report them descriptively, not as
evidence of partial generalisation.

Across the three configurations, the fraction of seeds that grok rises from $0/16$ to
$11/16$ to $16/16$. We interpret this only as a \emph{descriptive association across
three realized fraction$\times$split combinations}, not as a function of train fraction.
Each fraction uses one split seed ($1000$, $1001$, $1002$ respectively), and the three
training sets are not nested subsets of a single permutation. Train fraction and split
identity are therefore completely confounded: the observed quantities are the success
rates of three particular fraction-and-split combinations. Establishing a dependence on
fraction would require several split seeds for each fraction, preferably nested, with
hierarchical intervals. This observation is nevertheless orthogonal to the RSB
hypothesis that motivated the sweep.

\subsection{O1: overlap-distribution shape across the transition}
\label{sec:o1}

\texttt{frac40} is the only configuration statistically powered for overlap analysis
(16/16 replicas have both checkpoints). Before the RSB decision rule is evaluated,
however, the registered numerical-precision rail fires. Across the full population of
120 pairs in each checkpoint category, $100\%$ of pairs exceed the registered $10^{-6}$
fp32/fp64 tolerance. The mean discrepancy is $3.0\times 10^{-6}$ at the plateau and
$2.8\times 10^{-5}$ post-grokking. Under the registered rule, the formal O1 verdict for
\texttt{frac40} is therefore \reasoncode{}: an inconclusive result, not a null, and
reported as such.

We diagnosed the failed gate rather than discarding it. For the full population, we
compared the actual \emph{discrete alignment choice} under fp32 and fp64, not merely the
continuous overlap value. The comparison covers all 120 pairs in each of the 2 checkpoint
categories, for $240$ pair-category combinations. Within each category it checks
$15{,}360$ residual-stream and $61{,}440$ MLP permutation elements. There are
\emph{zero} mismatches in either category: fp32 and fp64 always select the same
alignment. The discrepancy is consistent with ordinary fp32 summation rounding in a
$227{,}313$-parameter normalized dot product. The expected scale is
$\approx \sqrt{N}\,\epsilon \approx 5.7\times10^{-5}$, the same order of magnitude as,
although not an exact match to, the observed mean $2.8\times10^{-5}$. Thus, the
discrepancy is consistent with numerical noise rather than alignment instability.

\paragraph{Why the precision rail could not have passed.} Two registered mechanisms
use this tolerance, but they are not the same rule. The \emph{numerical rail} excludes
any individual overlap value whose fp32/fp64 relative discrepancy exceeds $10^{-6}$ from
the \S7/\S8 statistics. The \emph{C0 reason code}, by contrast, is assigned to the whole
configuration only if this discrepancy exceeds $10^{-6}$ for more than $5\%$ of pairs.
The distinction does not alter the present result because $100\%$ of pairs exceed the
tolerance. The rule discussed below is the per-value rail, which requires fp32/fp64
agreement within $10^{-6}$ for every pair.

This tolerance is not commensurate with the arithmetic to which it is applied. The
quantity $q_{wt}$ is a normalized dot product over $N = 227{,}313$ parameters. In fp32,
where machine epsilon is $\epsilon \approx 1.19\times10^{-7}$, its accumulation has an
expected rounding scale of order $\sqrt{N}\,\epsilon \approx 5.7\times10^{-5}$---roughly
$57\times$ the permitted tolerance. This scale argument is heuristic, not a derivation
of the error: it assumes random-walk accumulation and does not represent the actual
summation order. With that qualification, the observed mean discrepancy of
$2.8\times10^{-5}$ is \emph{consistent with the expected fp32 error scale}. A $100\%$
failure rate at this threshold is consequently unsurprising and provides only weak
evidence that the overlap computation is unstable. It does not prove that the pipeline
is correct. The registered rule contains a defect that we did not anticipate, but that
defect does not license the separate claim that the \emph{alignment} is sound; as
Section~\ref{sec:method} shows, it is not.

A registered rule cannot be changed after its result is known. We therefore do not
loosen the original tolerance or overturn the registered outcome. Instead, we add a
supplementary criterion---fp32/fp64 permutation \emph{agreement}, rather than raw
overlap-value tolerance, for $\geq 95\%$ of pairs---and report it beside the original
result. \textbf{This criterion is post-hoc, as the git history shows exactly.} The body
of the pre-registration was committed on 2026-08-22 at 18:15. The \texttt{frac40}
overlap results were produced on 2026-08-23 at 02:59. The amendment introducing this
criterion was committed on 2026-08-23 at 06:54, \emph{after} both the data and the
registered \texttt{UNDETERMINED} outcome were known, and it was applied to the same
data. The heading of the pre-registration says so. Consequently, this criterion is
exploratory, not a second confirmatory pass.

\begin{figure}[h]
\centering
\includegraphics[width=0.9\textwidth]{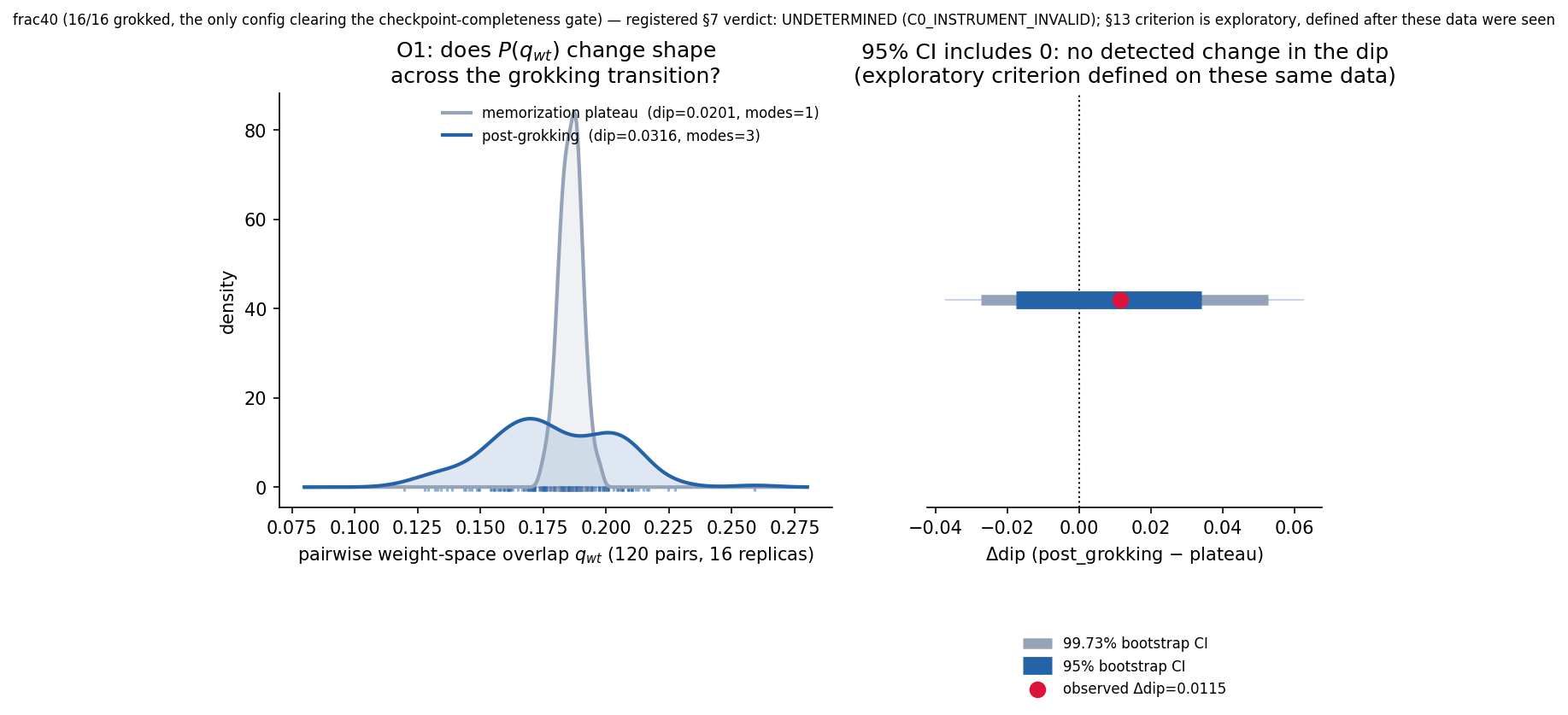}
\caption{Results for \texttt{frac40}, the only configuration that clears the
checkpoint-completeness gate. Left: KDEs of pairwise weight-space overlap $q_{wt}$ at the
memorization plateau (unimodal, dip $=0.0201$) and after grokking (3 KDE modes, dip
$=0.0316$); these are descriptive only. Right: bootstrap $\Delta$dip intervals. Only the
$95\%$ interval is interpreted, and it includes zero. The $99.73\%$ interval is shown for
reference but is not used as a decision rule (Section~\ref{sec:o1}). Registered outcome
(\S7 of the pre-registration):
\texttt{UNDETERMINED} (reason code: \reasoncode{}). Under the supplementary same-data
criterion (\S13 of the pre-registration, evaluated on all 120 observed pairs), no change
in the dip was detected.}
\label{fig:o1}
\end{figure}

Under the supplementary criterion, the 16/16-populated \texttt{frac40} bootstrap gives
an observed $\Delta\text{dip} = 0.012$ and a 95\% percentile-bootstrap interval of
$[-0.017, 0.034]$, which contains zero (Figure~\ref{fig:o1}). We report the 95\% interval
descriptively and do not apply the originally registered $99.73\%$ gate. That gate puts
each tail at the $0.135$th percentile. With $1{,}000$ resamples, this corresponds to
roughly the $1.35$th order statistic and is determined by one or two extreme resamples.
The reported $95\%$ interval places its tails near the $25$th order statistic of the same
$1{,}000$ resamples and is therefore much less sensitive to individual draws. Neither
interval has calibrated coverage, which is why we use the reported interval
descriptively rather than as a decision rule. The replica bootstrap also inserts $q=1$
whenever a replica is paired with itself---on average $\approx 7.3$ of $120$ dyads---whereas
the observed cross-replica overlaps are only $\approx 0.12$--$0.26$.

\textbf{What does this result show?} The post-hoc interval for the \emph{Hartigan dip
statistic} contains zero. Per D4, the test has no power at the simulated separations, so
this interval records a failure to measure, not a measured absence. It does not show that
the shape of the distribution is unchanged. The dip measures departure from unimodality
and is insensitive to location and scale. Across the same transition, the sample
standard deviation of $q_{wt}$ increases from $0.00444$ to $0.02471$, a factor of
$\approx 5.6$, and the post-transition kernel density estimate has three modes. Clearly,
there is detectable widening; we do not claim otherwise. Nor does an interval containing
zero establish absence. We specified neither an equivalence margin nor a smallest effect
size of interest, so this analysis excludes no RSB-consistent change of practical size.

One sanity check supports the mechanical prediction that motivated the choice of
$q_{wt}$ as the primary statistic: the function-space overlap $q_{fn}$ is at its ceiling
after grokking and is therefore uninformative. At post-grokking, $q_{fn}$ has mean
$=1.0000$ and sd $=0.0000$: every pair of grokked replicas agrees on exactly $100\%$ of
held-out predictions. Its mean at the plateau is $=0.023$. Thus, the pre-registered
rationale for replacing the metric is not merely plausible; in this sample, $q_{fn}$
reaches its post-grokking ceiling ($1.0000$).

\subsection{O2: ensemble-size diminishing returns}

Again, \texttt{frac40} is the only configuration that clears the
checkpoint-completeness gate for O2. The pre-registration divides its 16 replicas into
two disjoint groups of eight. Group A supplies the mode count $M_A$; group B supplies the
ensemble curve. The two halves of O2 are thus evaluated on different networks. The eight
group-A replicas yield $\binom{8}{2} = 28$ pairs, for which we reuse the aligned overlap
matrix.

For these 28 post-grokking pairs, the KDE mode count is $M_A = 1$ (mean $q_{wt} = 0.178$,
sd $= 0.021$). Its registered status matches that of O1:
\textbf{Registered outcome: \texttt{UNDETERMINED} (reason code: \reasoncode{})}, because
the same fp32/fp64 rail covers these pairs. Under the supplementary criterion, the rail
labels the result $M_A = 1$. We say ``labels,'' not ``validates,'' because a post-hoc
criterion (Section~\ref{sec:o1}) cannot confer validity.

The value $M_A = 1$ does not conflict with the three post-grokking modes in O1, and
neither result is evidence about the other. They use different samples: 28 pairs from
eight replicas here, compared with 120 pairs from all 16 replicas in O1. At this scale, a
mode count depends strongly on sample size. We report both results but do not treat them
as mutually corroborating.

\begin{figure}[h]
\centering
\includegraphics[width=0.55\textwidth]{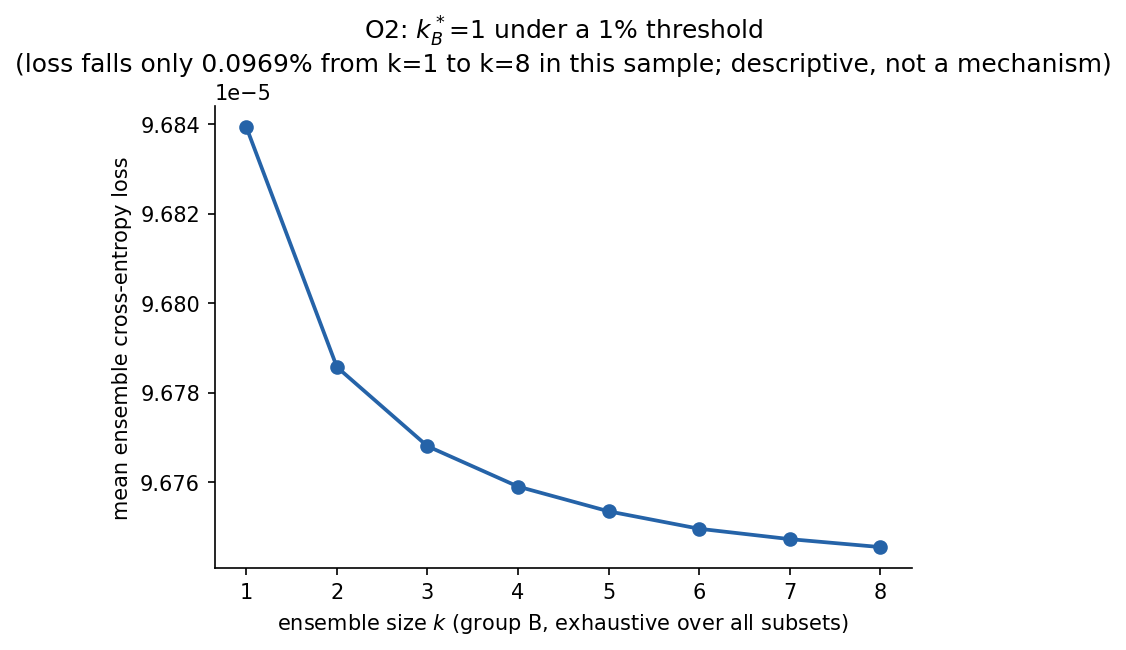}
\caption{Group-B ensemble cross-entropy loss against ensemble size $k$, exhaustively
enumerated over all subsets. The curve is not flat: mean loss at each $k$ decreases monotonically
from $9.6839\times10^{-5}$ at $k=1$ to $9.6746\times10^{-5}$ at $k=8$. This $0.0969\%$
decrease lies within the pre-specified $1\%$ threshold, giving $k^*_B=1$. We therefore
describe the curve as empirically near-flat in this sample and infer no mechanism from
it; Section~\ref{sec:o2disc} retracts the earlier interpretation as a forced floor
effect.}
\label{fig:o2}
\end{figure}

For group B, the smallest ensemble within 1\% of the full 8-member loss is $k^*_B = 1$.
The loss curve is not constant. Mean loss decreases monotonically from
$9.6839\times10^{-5}$ at $k=1$ to $9.6746\times10^{-5}$ at $k=8$, a relative decrease
of $0.0969\%$. The decrease simply lies within the pre-specified $1\%$ threshold
(Figure~\ref{fig:o2}). We therefore call the curve \emph{empirically near-flat in this
sample under a $1\%$ threshold}; we do not describe the result as mechanically forced.

\label{sec:o2disc}
In particular, the accuracy ceiling does not force $k^*_B = 1$. All 16
\texttt{frac40} replicas reach $\texttt{test\_acc} = 1.0000$ exactly
(\texttt{min}$=$\texttt{max}$=1.0$), but equal accuracy does not imply equal
cross-entropy. Perfect classifiers may differ in confidence, and probability averaging
may still reduce cross-entropy appreciably. Saturated accuracy makes small, similar
per-member losses plausible, but these data cannot establish that quantitative claim:
\texttt{o2\_frac40\_result.json} stores only the mean loss at each $k$, not the
individual member losses or their variance.

The measure faces two further limits. First, the O2 pipeline produced only one
$(M_A, k^*_B)$ point. One point cannot test whether ensemble returns track the modality
of $P(q)$, so the intended correlation is tested in neither direction. Second, a genuine
O2 test would require the individual member losses and their variance, together with the
registered 200-split companion (Section~\ref{sec:deviations}, D3). We leave these as
design lessons rather than adding a post-hoc patch.

\section{Discussion}

Taken as a whole, this study does not give a confirmatory null for its primary question.
The registered rule returns \texttt{UNDETERMINED}, and the audit in
Section~\ref{sec:deviations} shows that the instrument is invalid for that question. The
supported conclusion is narrower and exploratory. For the only configuration that
clears the checkpoint-completeness gate, a criterion defined on the same data detects no
change in the Hartigan dip statistic of the aligned overlap distribution, while the
standard deviation of that distribution increases by a factor of $\approx 5.6$ across
the same transition. The study neither tests nor claims to test whether RSB effects
appear \emph{transiently}, closer to the transition boundary.

\paragraph{A successor hypothesis, offered as exploratory.} Among the three grokking
configurations, \texttt{frac40} reaches its convergence criterion earliest and does so
for all $16/16$ replicas, whereas \texttt{frac30} lies at the borderline with $11/16$
(Section~\ref{sec:reliability}). The contrast suggests a natural next question: if an RSB
signature exists, is it concentrated near the reliability boundary rather than in a
regime that groks readily?

This hypothesis arises from gaps in the experimental design; it is \emph{not} a
prediction derived from spin-glass theory. An earlier draft argued that RSB effects in
canonical spin glasses are generally strongest near a critical point and weaker deep in
an ordered phase, and it used that argument to present the successor hypothesis as
physically motivated rather than as a post-hoc rescue. \textbf{We retract that argument.}
It is not a general property of canonical models. Indeed, the opposite holds at the
low-temperature end of the Sherrington--Kirkpatrick model. \citet{auffinger2017sk} prove
that, at zero temperature, the Parisi measure of the mixed $p$-spin model has infinitely
many points in its support. This establishes Parisi's prediction that the SK functional
order parameter is not a step function there and implies that \emph{the number of levels
of broken replica symmetry diverges as the temperature goes to zero}. In that model,
replica symmetry breaking therefore has \emph{more} levels, not fewer, deep in the
ordered phase. It cannot justify the successor hypothesis.

We offer no substitute theoretical claim. Establishing the near-boundary regime as the
right place to seek RSB structure would require a model-specific derivation absent from
this study. Testing the hypothesis would in any event require new training, because
\texttt{frac30} fails the C1 gate and post-hoc power boosting is precisely what the
registration prevents.

Nor do we interpret O2 in either direction. Instead, we retain it as a design lesson: a
measure selected to avoid one confound---cross-entropy loss in place of $q_{fn}$ and its
accuracy ceiling---can meet a similar ceiling one derivative away after accuracy itself
saturates.

\subsection*{Limitations}
(1) \textbf{The instrument is invalid for the registered question.} Because the
alignment implementation does not preserve the network function
(Section~\ref{sec:method}), it does not actually quotient out the permutation symmetry on
which the analysis depends. Every overlap value in this paper inherits this defect.
(2) The registered numerical-precision rail does not clear for \texttt{frac40}; the
primary rule therefore returns \texttt{UNDETERMINED}. The dip result uses a criterion
defined after inspection of these data and is exploratory rather than confirmatory.
(3) No power calibration preceded the analysis. We therefore claim neither that any
configuration is adequately powered nor that a minimum detectable effect was
pre-specified. The post-hoc calibration in D4 gives the dip statistic no power at the
simulated separations, so its interval is uninformative rather than null-supporting. Only
one of the three train-fraction configurations clears even the checkpoint-completeness
gate; \texttt{nogrok} does not participate in that gate. The RSB question is untested,
not falsified, in the other two configurations. (4) Several registered controls did not
run (Section~\ref{sec:deviations}, D3). In particular, \texttt{nogrok} does not perform
its registered role of excluding continued optimisation and weight decay as causes of
change in $P(q)$. (5) The dip statistic tests unimodality only, and no equivalence margin
was specified. Thus, the study cannot exclude an RSB-consistent change in width, support,
moments, or ultrametric structure. (6) O2 supplies one descriptive point and tests no
correlation. (7) The study considers one small algorithmic task---modular addition with
$p=113$---one pinned architecture, and one split seed per configuration. It supports no
claim about larger scales or other task families.

\section{Successor hypothesis (not pursued in this study)}

An originally sketched, wider train-fraction sweep would fix neither the invalid
alignment instrument nor the absent pre-analysis power calibration. We therefore do not
spend further compute on it. Instead, we register a narrower hypothesis for future work:
RSB effects may be a near-transition-\emph{boundary} phenomenon. Testing this possibility
requires 2--4 new \texttt{train\_frac} configurations clustered near \texttt{frac30}
(e.g.\ \texttt{frac28}--\texttt{frac33}) to locate the boundary with statistical power,
together with an earlier, not-fully-converged checkpoint for an O2-style ensemble test
that is not limited by a floor. This hypothesis is open, dated, and recorded in the
study's catalog entry rather than pursued here, consistent with the project's practice
of not following every new question within the study that generated it.

\section{Conclusion}

The registered test of a change in replica overlap across grokking gives
\textbf{registered outcome: \texttt{UNDETERMINED} (reason code: \reasoncode{})}. It is
the non-function-preserving alignment, together with the failed precision requirement
and the protocol deviations in Section~\ref{sec:deviations}, that prevents confirmatory
inference. For \texttt{frac40}, a post-hoc interval calculated from the same data contains
zero for the Hartigan dip. Because the test has no power at the simulated separations
(D4), that interval is uninformative. Yet the overlap standard deviation increases by a
factor of about $5.6$. Meanwhile, the grokking rates across the three fraction-by-split
combinations remain confounded by split identity.

A valid test requires function-preserving alignment, matched-epoch controls, and
pre-specified power calibration. Even these corrections are not sufficient on their own.
Discrete permutation alignment leaves the ReLU positive-rescaling symmetry and the
within-head $Q/K$ and $V/O$ basis freedom intact, so a gauge-invariant order parameter
must also handle those degrees of freedom. We report this pilot because its instrument
failed in a locatable way and thereby identifies what the next attempt must fix.

\bibliographystyle{plainnat}
\bibliography{../refs/refs}

\end{document}